# Transferring learned patterns from ground-based field imagery to predict UAV-based imagery for crop and weed semantic segmentation in precision crop farming

Junfeng Gao[1,2], Wenzhi Liao[3,4], David Nuyttens[5], Peter Lootens[5], Erik Alexandersson[6], Jan Pieters[7]

1, Lincoln Agri-Robotics (LAR), Lincoln Institute for Agri-Food Technology, University of Lincoln, Lincoln, UK

2, Lincoln Centre for Autonomous systems (LCAS), University of Lincoln, Lincoln, UK

3, Flanders Make, Kortrijk, Belgium

4, Department of Telecommunications and Information Processing, Ghent University, Ghent, Belgium

5, Flanders Research Institute for Agriculture, Fisheries and Food (ILVO), Merelbeke, Belgium

6, Department of Plant Protection Biology, Swedish University of Agricultural Sciences, Alnarp, Sweden

7, Biosystems Engineering Group, Ghent University, Ghent, Belgium

**Corresponding author:**

**Email address: jugao@lincoln.ac.uk (J. Gao), wenzhi.liao@flandersmake.be (W. Liao).**

**Abstract:** Weed and crop segmentation is becoming an increasingly integral part of precision farming that leverages the current computer vision and deep learning technologies. Research has been extensively carried out based on images captured with a camera from various platforms. Unmanned aerial vehicles (UAVs) and ground-based vehicles including agricultural robots are the two popular platforms for data collection in fields. They all contribute to site-specific weed management (SSWM) to maintain crop yield. Currently, the data from these two platforms is processed separately, though sharing the same semantic objects (weed and crop). In our paper, we have developed a deep convolutional network that enables to predict both field and aerial images from UAVs for weed segmentation and mapping with only field images provided in the training phase. The network learning process is visualized by feature maps at shallow and deep layers. The results show that the mean intersection of union (IOU) values of the segmentation for the crop (maize), weeds, and soil background in the developed model for the field dataset are 0.744, 0.577, 0.979, respectively, and the performance of aerial images from an UAV with the same model, the IOU values of the segmentation for the crop (maize), weeds and soil background are 0.596, 0.407, and 0.875, respectively. To estimate the effect on the use of plant protection agents, we quantify the relationship between herbicide spraying saving rate and grid size (spraying resolution) based on the predicted weed map. The spraying saving rate is up to 90% when the spraying resolution is at $1.78 \times 1.78$ cm$^2$. The study shows that the developed deep convolutional neural network could be used to classify weeds from both field and aerial images and delivers satisfactory results.

1. **Introduction**

Weeds generally emerge in natural fields. They cause a tremendous loss in crop yield and quality (Oerke, 2006). To sustain an increasing worldwide population with sufficient crop production with limited land resources, new smart farming approaches are highly needed to increase yield while minimizing environmental impacts. Precision crop farming (PCF) is a whole-farm management approach using information technology, satellite position data, remote and proximal sensing data combined in smart decision tools, and precision application techniques. The main objective of PCF for weeding is to allocate the right doses of input at the right time and in the right position for improving crop performance and environmental sustainability (Zhang et al., 2002).

Conventional weed managements rely on whole-field management (Shaw, 2005) without considering spatial differences in weed distribution. Site-specific weed management (SSWM), a subfield of PCF, refers to the spatially variable application of a weed management strategy without spraying chemicals uniformly (Burgos-artizzu et al., 2011). It allows appropriate use of chemicals and thereby potential reduction of adverse impacts to the environments. However, one of the main challenges to achieve SSWM is developing a reliable and accurate weed recognition system under unstructured environments such as field conditions. To address this challenge, a few studies have been attempted, mainly using computer vision with RGB images (Gao et al., 2020), multispectral and hyperspectral images(Gao et al., 2018b; Zhang et al., 2019), and spectroscopy in visible and near-infrared regions (Shapira et al., 2013). In addition, machine learning has become an increasingly popular approach for precise real-time weed and crop detection in fields (Wang et al., 2019). A typical processing procedure based on machine learning includes pre-processing, feature extraction, classification or recognition. While this learning-based approach has made further improvements for weed recognition compared to a system developed only based on image processing, its effectiveness highly relies on robust feature extraction which requires careful hand-crafted designs and significant domain expertise. By contrast, deep learning, a subfield of machine learning, enables hierarchical learning features and discovering potential complex patterns automatically from large datasets (LeCun et al., 2015). It has shown impressive advancements in solving various complex problems. In the agricultural domain, deep learning is also a promising technique with growing popularity. Kamilaris and Prenafeta-Boldú (2018) carried out a survey about deep learning in agricultural applications. It indicates that deep learning provides higher accuracy, outperforming typically machine learning techniques, in weed detection(Gao, 2020; Gao et al., 2020; Milioto et al., 2017), crop disease recognition (Gao et al., 2021; van de Vijver et al., 2020) and plant stress phenotyping (Singh et al., 2018).

Generally, weed recognition with deep learning can be categorized into two groups: ground-based and aerial remote or proximal sensing. Satellite imagery could be utilized for monitoring weed invasion at a large scale, but it is difficult to map weed individuals due to low spatial resolution (Peerbhay et al., 2016). In terms of ground-based systems, Wang et al., (2007) developed a real-time, embedded weed detection system based on a tractor platform with an integrated sprayer. Furthermore, mobile robots have become widely utilized for in-field scenarios to perform difficult tasks while maintaining high-level accuracy and robustness (Emmi and Gonzalez-de-Santos, 2017). The use of ground-based autonomous robots can decrease the human labour input required for weed management. Typically, a vision-based system in combination with deep learning is one of the main components in field robots (Bawden et al., 2017; Lottes et al., 2018). A field robotic platform is also commonly used to collect images, providing a large number of datasets for deep learning research. Recently, unmanned aerial vehicles (UAVs) are becoming an increasingly popular tool for various remote sensing applications. Compared to ground-based robotic platforms, UAVs have been considered more efficient as they allow fast data acquisition with relatively high resolution and low cost (Torres-Sánchez et al., 2015). Perez-Ortiz et al., (2016) discussed patterns and features for inter- and intra-row weed mapping using UAV imagery. Gao et al., (2018a) fused pixel- and object-based features from aerial images with a random forest classifier for weed mapping in a maize field. Furthermore, Dian Bah et al., (2018) explored unsupervised data labelling obtained from inter-row weeds for weed mapping on drone-based aerial images. In addition, Sa et al., (2018a, 2018b) developed an end-to-end semantic weed mapping framework based on aerial multispectral images and deep learning.

Ground-based images can be collected with a ground-based vehicle. The collected images are typically used for developing weed recognition systems in ground-based platforms such as autonomous weeding robots and spot spraying tractors. UAVs enable mapping weeds with a global view based on orthophotos generated from collected overlapping images. Data from these two platforms (UAV-based, ground-based) is typically processed independently for weed detection. This is because the two data sources have their characteristics in terms of resolution and scale. For example, the training samples (tiles or regions) for aerial weed mapping with deep learning, are usually randomly cropped from an orthomosaic map generated by aligning overlapped aerial images. They are typically in low resolution with blurry object boundaries, or they might have artifacts from the process of image alignment to produce orthomosaic imagery. By contrast, samples for field weed detection are captured in a closer overhead view, and therefore are having a higher image resolution. The resolution difference also impacts the process of annotation for training networks. The annotation of the cropped small tiles (low-resolution) usually consumes more time and is error-prone compared to ground-based proximal images (high-resolution) due to spatial resolution differences.

Despite of sharing a same task for weed detection, these differences bring the challenge of developing and training one network architecture to achieve a satisfactory performance in the two domains (field and aerial images). Bridging the domain gap from ground-based images to aerial images for weed mapping is crucial for exploiting two different datasets, transferring knowledge learned between the two similar tasks. To the best of our knowledge, no specific studies have attempted to connect or utilise these two different datasets for weed mapping.

In this study, we proposed a pipeline to pre-process field images in order to develop an effective deep convolutional neural network with a weighted loss function for weed and crop semantic segmentation in precision farming. Specifically, our model was developed based on field images alone and then applied to predict unseen tiles from the orthomosaic imagery with larger mapping areas but at lower spatial resolution. Then the weed map was generated from the tile predictions. Furthermore, different models were compared, and the optimal class weight was determined for weed and crop segmentation. Moreover, feature maps from different layers in the developed model were displayed for its accountability and interpretability.

The main contributions of our work are summarized as follows:

(1) It bridges the gap between using ground-based field images (high-resolution data) and UAV images (low-resolution data) for weed and crop semantic segmentation which has not been fully exploited so far. This cross-domain learning allows for the generation of weed maps with UAV-based remote sensing data by using different source datasets for prediction performance by bootstrapping.
(2) It develops a deep convolution neural network that integrates a custom image pre-processing pipeline to remedy the influence of the differences between the two domains for predictions.
(3) It systematically investigates and determines the relationships between the grid size (spraying resolution) based on the predicted weed map and herbicide spraying saving rate for the spot spraying application.
(4) It provides a comprehensive analysis to discuss what features were learned by the model based on layer-wise visualization. This analysis contributes to the development of deep learning and remote sensing in precision farming tasks.

## 2. Materials and Methods

### 2.1, Image dataset collections

#### 2.1.1 Field image collection

The field images were manually collected with a visual camera (Nikon D7200) under different lighting conditions from two maize fields near Ghent and two maize fields near Merelbeke in Belgium in 2018. The growth stages of maize plants range from 2 unfolded leaves to 5 unfolded leaves. The distance between the camera and the ground was around 1 m which is not strictly fixed in order to create more variations in the images. Multiple weed species were found in these fields. In our study, we only classified three classes

(weed, maize, soil) regardless of different weed species. More than 10000 images (48G) were randomly collected during the early growth stages of maize crop in these four maize fields.

**2.1.2 UAV-based image collection**

The UAV-based image data was collected at the experimental fields of the Flanders Research Institute for Agricultural, Fisheries, and Food Research (ILVO) in the agricultural region of Merelbeke, East-Flanders Province, Belgium in 2016. Several weed species like bindweeds (*Convolvulus*), lamb's quarters (*Chenopodium album*) and crabgrass (*Digitaria sanguinalis*) species were naturally infested in the field plots. A 12 coaxial rotors UAV (Hydra-12 Onyxstar, Mikrokopter, Germany), equipped with a lightweight visual camera (Sony Alpha 6000, Sony, Japan), was employed to collect aerial imagery at an altitude of 20m above ground. The details of the field experiments and imaging parameters can be seen in Gao et al., (2018a). The two cameras' technical specifications and parameter settings are listed in Table 1. The UAV flew over the maize experimental field every week from the emergency of maize plants until the maize canopy covered the crop rows.

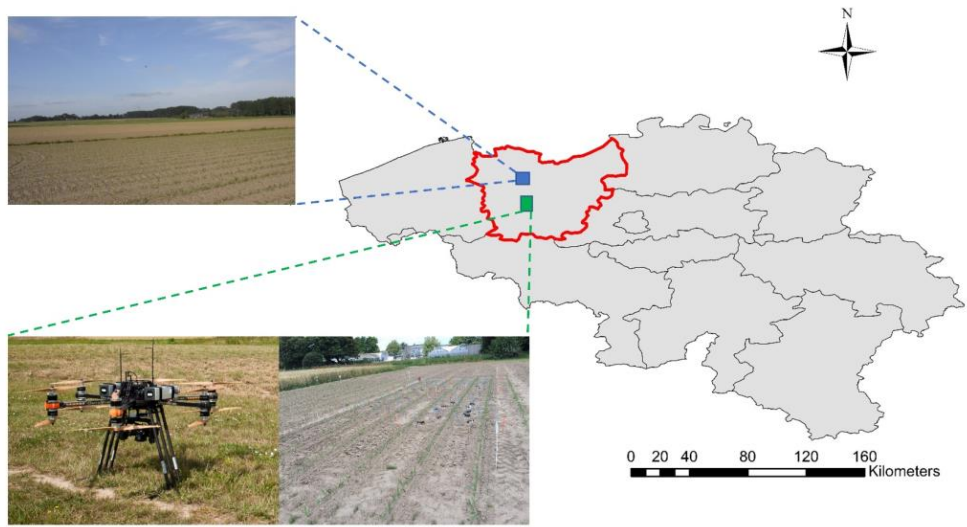

**Figure 1. Field and UAV experiments data collection location**

**Table 1. Camera types and settings in two experiments**

|  | Ground-based monitoring | UAV-based monitoring |
|---|---|---|
| Camera | Nikon D7200 | Sony Alpha 6000 |
| Resolution | 6000 × 4000 pixels | 6000 × 4000 pixels |
| Lens type | AF-S-NIKKOR 18-140mm | Sony E 35mm F 1.8 OSS |
| Shutter time | 1/800 s | 1/2000 s |
| ISO | 1600 | 320 |
| GSD | ___ | 1.78 mm/pixel |

## 2.2, Overall workflow

Figure 2 depicts the entire pipeline implemented in this paper. There are two main components in the pipeline. One is training the network using field images, and another is using the developed network to predict a region of interest (ROI) from orthomosaic images. For the training dataset, we did not directly use original field images (6000*4000) because of the processing difficulties associated with memory management in available GPUs. We instead first resized original images and then randomly cropped image tiles (512*512) from the resized images (1200*800). There are 5000 field image tiles (512*512) in total which were used to train the deep convolutional neural network. For the validation dataset, 621 field images were used. After the network was trained, not only its performance was evaluated in the field test dataset, but also its generalizability was validated by directly testing UAV-based imagery which had never been seen before by the network. In terms of orthomosaic image generation, the Agisoft PhotoScan v1.2.3 (Agisoft LLC, St. Petersburg, Russia) software was employed. The steps to process aerial images in Agrisoft PhotoScan include image alignment, dense cloud building, 3D mesh building, texture building, digital elevation model building and finally orthomosaic photo generation (blending mode). We set high accuracy and 4000 key points limited for image alignment(Gao et al., 2018a). The other parameters were set by default in PhotoScan.

It is critical to apply the pre-processing for field images to close the gap between field images and aerial images to the utmost. The two types of datasets share the same object classes (soil, crop, weed), but they represent the main differences in object scales, resolution. Intuitively, one straightforward approach is to downsample and then apply Gaussian smoothing in the field images. In addition, Gamma correction (power law transform) was performed on each input image. All these operations were done before the images were fed into networks for training.

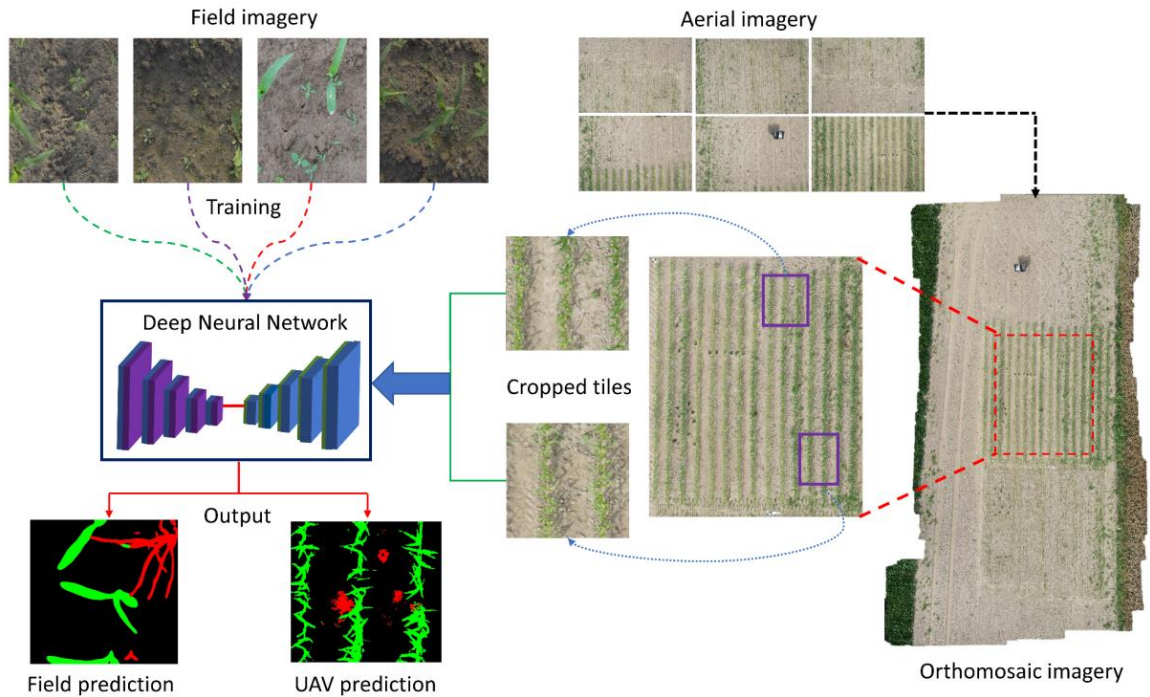

Figure 2. The diagram of this study. The network was trained with field images alone and predicted both field images and cropped aerial images to map weeds in fields.

**2.3, Encoder-decoder network architecture for semantic segmentation**

Fully convolutional networks (FCN) are suitable for field and remote sensing imagery and have demonstrated promising results in the studies for semantic segmentation of very high-resolution UAV imagery(Huang et al., 2019; Sa et al., 2018a). FCN models, such as FCN32s, FCN-16s and FCN-8s (Long et al., 2015), DeconvNet (Noh et al., 2015), and SegNet (Badrinarayanan et al., 2017), generally have an encoder-decoder network architecture. The encoder comprises a series of convolutional operations, activation and pooling operations. Semantic features from low-level to high-level could be extracted at the end of the encoder process. Because of pooling layers, the spatial resolution of the encoder output is smaller than the original input images. By contrast, the decoder part upsamples the features learned from the encoder part and generates the final prediction results with the same spatial size as the original input images. The upsampling layers, such as deconvolution layer, in the decoder are also capable of learning mapping from feature maps to semantic segmentation results. The proposed network, illustrated in Fig. 3, also has the encoder-decoder symmetric architecture. There are 39 layers in total with 29 convolutional layers (14 for encode part, 15 for decode part), 5 max pooling layers and 5 upsampling layers. Each convolutional layer (1-28) is followed by batch normalization and rectified linear unit (ReLu) activation operations. The last convolutional layer, however, employs "Softmax" as activation function. The size of all the kernels in the convolutional layers is 3×3. The indices (the location of the maximum feature value in every max pooling operation) are recorded during the max pooling (2×2) layers and then are passed to the corresponding upsampling layers (2×2), which is similar to SegNet (Badrinarayanan et al., 2017). In this way, substantial spatial information is saved for more accurate object boundary prediction. Furthermore, we use skip connections to concatenate former layers together with upsampling layers in the decoder part. This could fuse the fine-grained features from former layers, which is different from the SegNet using sum operations instead. There are 42,037,391 parameters in total with 42,018,441 trainable parameters and 18,950 non-trainable parameters. The number of parameters is higher than the other networks (FCNs, UNet and SegNet). The data augmentation was used and the loss values in the training and validation datasets were monitored to determine the timing to stop training further for preventing the risk of overfitting.

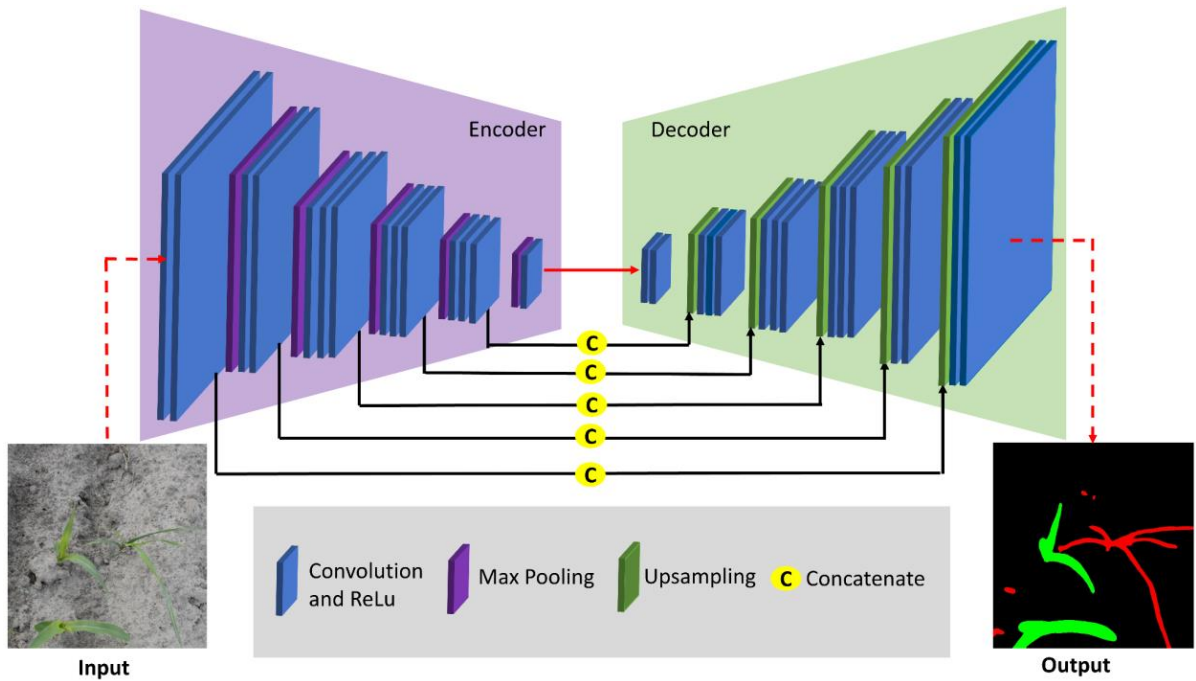

Figure 3. The developed encoder and decoder based neural network architecture.

**2.4, Weighted loss**

Designing the loss function is key for training a robust and high-performance network. The most commonly used loss function for semantic segmentation is pixel-wise cross-entropy loss. In our study, the frequency of appearance for each class (weed, maize and soil background) is heavily imbalanced. Generally, the number of soil background pixels is far more than pixels of crop and weed in field images at early growth stages of the crop. Just using the standard loss function without adaption would make a deep neural network model tend to only correctly classify dominant class pixels (soil background), ignoring the importance of weed and crop pixels. In order to draw the same attention in different classes for network training, a weighted loss function was finally used in the training process.

$$L = -\frac{1}{N}\sum_{i=1}^{N}\sum_{c=1}^{C} W_c * y_{o,c} \log p_{o,c}$$

where $N$ is the number of observations, $C$ is the number of classes (weed, crop and soil), $W_c$ is the weight for class $c$, $y$ is the binary indicator (0 or 1) if class label is corrected classified for observation $o$, p is predicted probability observation $o$ being of class $c$. we quantified the weights for each class by the following calculation.

$$W_c = \log \frac{\max(P_i, \ i \in \{1, \dots C\})}{P_c}$$

where $P_c$ is the number of pixels belonging to class $c$. Based on the statistics of pixels in the training dataset, the weights for weed, crop and soil background classes are 2.5, 1.5, and 1, respectively.

## 2.5 Post-processing with fully connected conditional random fields (CRFs)

CRF has traditionally been used to smooth noisy segmentation edges (Kohli et al., 2009), typically coupling nodes in the neighborhood favoring the same label assignment to spatially proximal pixels. However, these short-range of CRFs is to process the spurious segmentation results of weak classifiers built on top of local hand-crafted features (Chen et al., 2018). In order to overcome this limitation, fully connected CRFs are designed to connect to a deep convolutional neural network. By combining single pixel prediction and shared structure through unary and pairwise terms, the fully connected CRFs establish pairwise potential on all pairs of pixels in an image (Krähenbühl and Koltun, 2011). The energy function for a model is

$$E(x) = \sum_i \theta_i(x_i) + \sum_{ij} \theta_{ij}(x_i, x_j)$$

where $x$ stands for label assignment for each pixel, $\theta_i(x_i)$ represents pixel-wise unary likelihood which equal to $-logP(x_i)$ where $P(x_i)$ is label assignment probability at pixel $i$ calculated by the model. $\theta_{ij}(x_i, x_j)$ is the unary potential, given in below.

Efficient interference is obtained by establishing pairwise potential while employing a fully connected graph such as connecting all pairs of image pixels $i, j$. A linear combination of Gaussian kernels can be used to qualify the pairwise edge potential (Krähenbühl and Koltun, 2011).

$$\theta_{i,j}(x_i, x_j) = \mu(x_i, x_j) \left[ w_1 \exp\left( \frac{-|p_i - p_j|^2}{2\sigma_\alpha^2} - \frac{|q_i - q_j|^2}{2\sigma_\beta^2} \right) + w_2 \exp\left( \frac{-|p_i - p_j|^2}{2\sigma_\rho^2} \right) \right]$$

$$\mu(x_i, x_j) = \begin{cases} 1 & if\ x_i \neq x_j \\ 0 & otherwise \end{cases}$$

where µ is a label compatibility function and the remaining expression is a Gaussian kernel depending on the feature extracted from pixels $i$ and $j$, and is weighted by $w_1, w_2$. The kernel can be divided into two components. The first bilateral term is called appearance kernel and depends on both pixel positions denoted as $p$ and RGB colour denoted as $q$. This term could convert pixels with similar colour and position to have the same labels. The second kernel called the smooth kernel only depends on pixel positions, removing small isolated pixels or regions. The scales of Gaussian kernels are defined by $\sigma_\alpha, \sigma_\beta, \sigma_\rho$. Parameters $\sigma_\alpha, \sigma_\beta$ control the degree of nearness and similarity (Shotton et al., 2009).

### 2.6. Evaluation metrics

Two common metrics, mean pixel accuracy and mean intersection over union (mean IoU), are used to evaluate pixel-wise semantic segmentation (Thoma, 2016). Mean pixel accuracy provides information about the overall effectiveness of the classifier. It is the ratio of the correctly predicted pixel numbers to the total pixel numbers in the entire dataset. Mean IoU, also called Jaccard similarity coefficient, compares the similarity and diversity of the complete sample set. The two metrics are defined in the following Equations.

$$\frac{1}{m} \sum_{i=1}^{m} \frac{n_{ii}}{t_i} \tag{5}$$

$$\frac{1}{m} \sum_{i=1}^{m} \frac{n_{ii}}{t_i + \sum_{j=1}^{m} n_{ji} - n_{ii}} \tag{6}$$

where $t_i$ is the total number of pixels of class $i$ in the ground truth image, $n_{ji}$ is the number of pixels of class $j$ predicted as belonging to class $i$ and m is the total number of classes.

### 2.7 Method implementation

Data augmentation was employed in the training dataset at the stage of network training. All networks were implemented using the Tensorflow framework and were trained with four NVIDIA Tesla P100-SXM2 GPU (16GB memory). The network was trained with the Adam optimizer (Kingma and Ba, 2015) using the initial learning rate of 0.005. In order to achieve the optimal coverage point, the learning rate was dropped by 0.1 to maintain further learning once the loss value stays the same in 10 consecutive batches. The exponential decay rates for the first- and second-moment estimates were set as 0.9 and 0.999, respectively. The batch size was set as 24. All the other parameters were initialized using the approaches provided by He et al., (2015).

## 3. Results

**Field-to-field prediction**

In our experiment, three types of semantic classes (soil, weed, and maize) were investigated with the deep neural network-based pixel-wise classifications. Table 2 presents the detailed metrics of the three classes and the overall performances of the deep neural networks. In the FCN network family, we can see the FCN-8s (OA=0.797, mIOU=0.736) provided better predictions compared to FCN-16s (OA=0.768, mIOU=0.704) and FCN-32s (OA=0.749, mIOU=0.680), indicating that increasing the number of upsampling or unpooling layers is one of the effective ways to achieve finer predictions. The approach of Long et al., (2015) has similar conclusions. The proposed network performed best among other networks with 0.859 mOA and 0.767 mIOU. Interestingly, UNet obtained higher mOA (0.820) compared to SegNet (mOA=0.811). However, its mIOU value (0.698) is relatively lower than SegNet (0.750). In terms of predictions in different classes, it is obvious that soil background class is much more effortless to distinguish for networks due to its large colour differences compared to plant classes. All the networks achieved above 0.970 IOU values for the soil background class. Weeds were more difficult to recognise compared to the maize crop. This is mainly because maize crop tends to have similar colour and morphological characteristics while weeds generally present larger differences in terms of size, growth stage and morphology as multiple weed species appeared in our study fields. Thus, the network learns well maize crop patterns, leading to better performance in classification. Figure 4 displays the prediction of the examples from the test field dataset. The proposed network is able to detect small-sized weeds and produces a precise segmentation boundary. Moreover, it also performed well in high occlusion examples such as #2 (occlusions between weed and crop) and #3 images (occlusions between crop and crop). These are generally quite challenging images to segment well by using conventional image processing algorithms. From the semantic classification results, we concluded that the proposed network is very capable of weed and crop classification in the early growth stage under field conditions, despite the heterogeneous weed patterns.

**Table 2. Performance of the deep neural networks in the field test dataset**

| Network | Soil background | | Weed | | Crop | | Overall | |
|---|---|---|---|---|---|---|---|---|
| | OA | IOU | OA | IOU | OA | IOU | mOA | mIOU |
| FCN-32s | 0.990 | 0.972 | 0.439 | 0.371 | 0.819 | 0.698 | 0.74.9 | 0.680 |
| FCN-16s | 0.992 | 0.973 | 0.514 | 0.435 | 0.799 | 0.705 | 0.768 | 0.704 |
| FCN-8s | **0.993** | 0.977 | 0.595 | 0.514 | 0.803 | 0.716 | 0.797 | 0.736 |
| UNet | 0.987 | 0.977 | 0.689 | 0.415 | 0.784 | 0.703 | 0.820 | 0.698 |
| SegNet | **0.993** | **0.979** | 0.634 | 0.534 | 0.807 | 0.736 | 0.811 | 0.750 |
| Proposed | 0.989 | **0.979** | **0.746** | **0.577** | **0.841** | **0.744** | **0.859** | **0.767** |

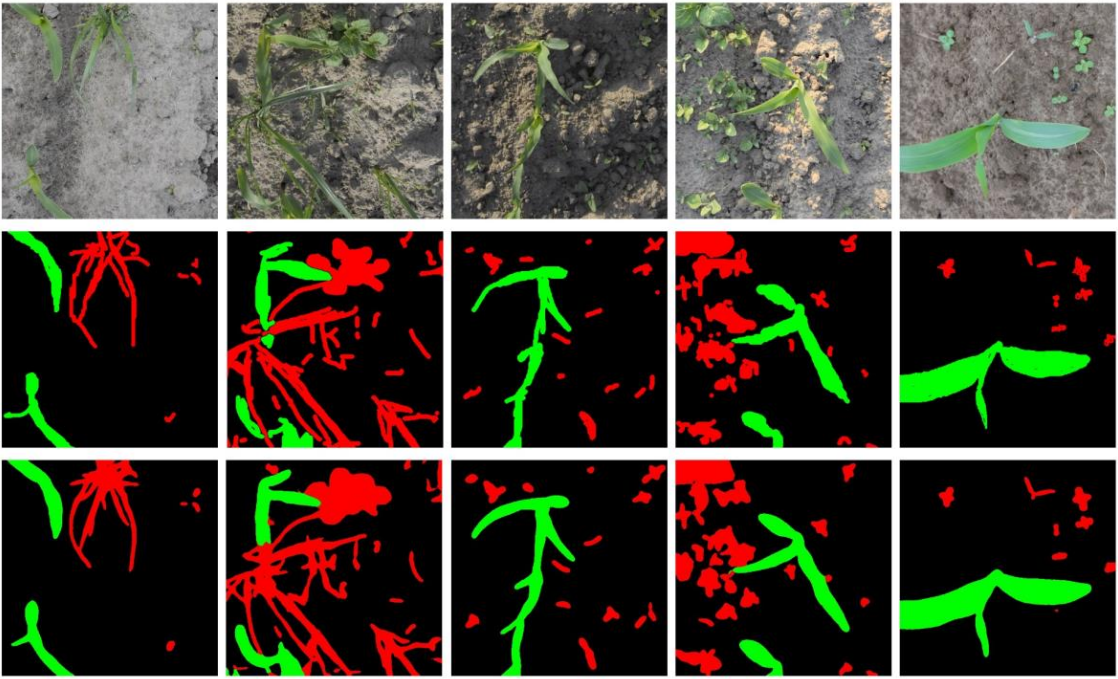

Figure 4. The semantic segmentation of field examples in the test dataset. The first row contains original field images. The second row is ground truth, and the third row is prediction results by the proposed network.

**UAV imagery prediction**

In this study, we want to produce a network that could predict both aerial images from a UAV platform and field images. To make it work, preprocessing was firstly applied to close the differences between field images and aerial images. Figure 5 displays the selected maize field from UAV orthomosaic imagery and its ground truth map (GTM). Note that this maize field is from a different field where the field images were collected, but the maize growth stages are similar (3-4 true full leaves). Therefore, the robustness and generalization of the network can be tested as well. It can be seen that maize plants are heavily overlapping in the vertical direction. Weeds are distributed both in the intra- and inter-rows. The detailed results of the networks in this dataset are reported in Table 3. It is observed that FCN-8s performed better than FCN-32s and FCN-16s which is similar to the field dataset discussed above. The proposed network (0.617), UNet (0.601) and SegNet (0.609) achieved better mIOU values than the FCN family with the best mIOU (0.538) achieved by FCN8. The proposed network works best among other networks. However, compared to performance in the field dataset, all the corresponding metrics in the networks decreased. This is mainly because the networks were trained with field images alone. Although the preprocessing algorithms were applied, trying to close the gap between the two datasets, the resolution and object (weed species) differences were still present. The weed/crop maps predicted by some of the networks are illustrated in Fig. 6. All the networks are able to detect inter- and intra-row weeds. The FCN16 has relatively rough prediction results, and weeds are overestimated compared to GTM. The proposed network refined the predictions and achieved 0.617 mIOU, though it still classified more weed pixels from other class pixels. The over-estimated weeds could be acceptable in practice as farmers generally prefer a conservative option to treat a larger area than really needed to ensure a better crop development (Borra-Serrano et al., 2015).

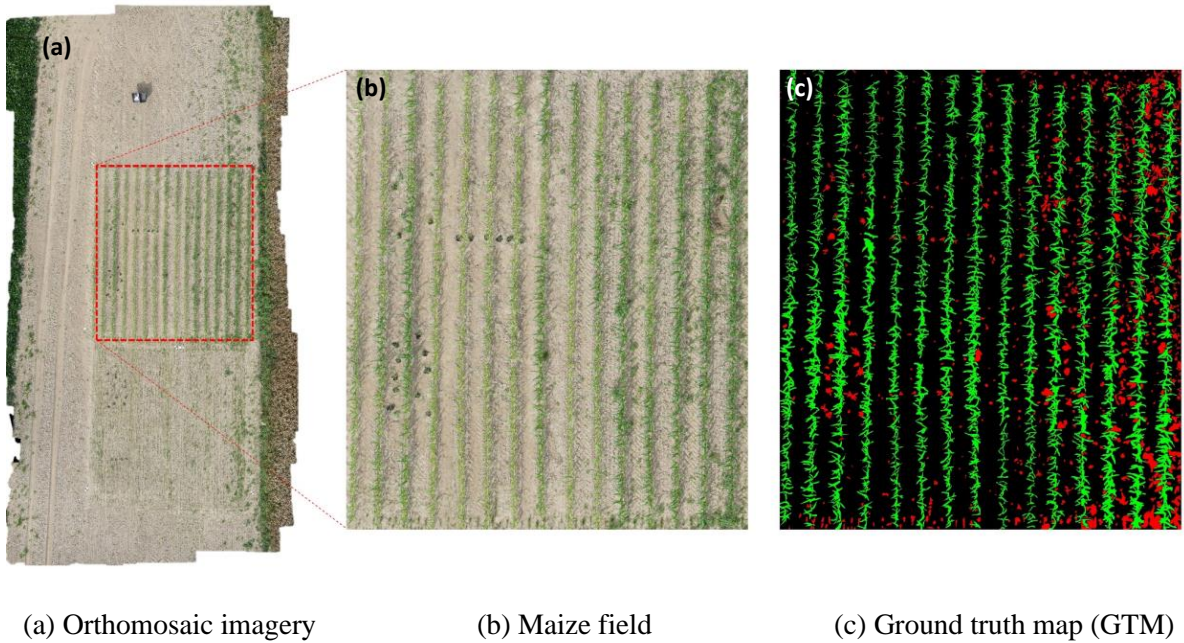

(a) Orthomosaic imagery          (b) Maize field          (c) Ground truth map (GTM)

Figure 5. The remotely sensed maize field from the UAV platform; (a) orthomosaic imagery, (b) the cropped interested field for testing performances of the networks, (c) manually labelled ground truth map (GTM).

Table 3. Performance of the deep neural networks in the UAV orthomosaic imagery

|         | Soil background | | Weed | | Crop | | Overall | |
|---------|------|------|------|------|------|------|------|------|
| Network | OA | IOU | OA | IOU | OA | IOU | mOA | mIOU |
| FCN-32s | 0.760 | 0.562 | 0.389 | 0.169 | 0.483 | 0.328 | 0.544 | 0.353 |
| FCN-16s | 0.884 | 0.623 | 0.397 | 0.205 | 0.557 | 0.375 | 0.613 | 0.401 |
| FCN-8s | 0.903 | 0.767 | 0.525 | 0.356 | 0.608 | 0.491 | 0.679 | 0.538 |
| UNet | 0.935 | 0.829 | 0.534 | 0.402 | 0.647 | 0.573 | 0.705 | 0.601 |
| SegNet | 0.913 | 0.811 | 0.569 | 0.434 | 0.667 | 0.581 | 0.715 | 0.609 |
| Proposed | 0.932 | 0.815 | **0.582** | **0.441** | **0.689** | **0.596** | **0.734** | **0.617** |

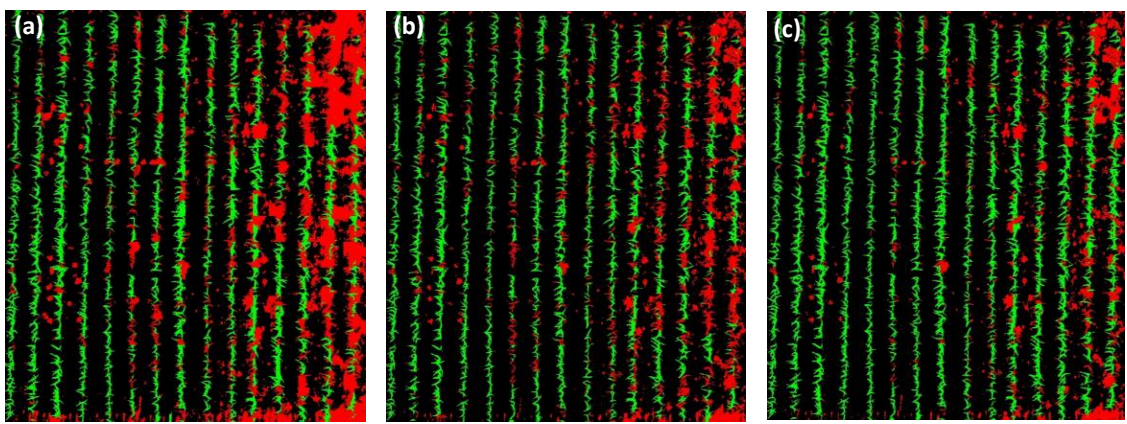

Figure 6. The maps predicted from the FCN16 (a), SegNet (b) and the proposed network (c).

Based on the predicted weed map from the proposed network, the weed stress is visualised in Fig.7(a). The corresponding spraying prescription map is displayed in Fig.7(b). The prescription weed map is a map that could be used to control nozzles of a spraying boom for spot spraying. In spot spraying, each nozzle is controlled individually instead of broadcast spraying by switching on all nozzles. In this case, the ground area in each grid of the prescription map (Fig.7(b)) could be treated differently based on the presence of pixels within that grid. From Table 4, we can find that 238 out of 504 grids (the total number of grids) are free of weeds, saving 47% herbicides compared to conventional broadcast spraying. It would save more with the increase of spraying resolution. As illustrated in Table 4, the herbicide saving rate could increase up to 90.9% if 10×10 spraying resolution (1.78×1.78 $cm^2$) is guaranteed. The spraying rates based on the prescription map at all three grid sizes are slightly higher than those from the GTM. This also indicates the proposed model predicted more weed pixels than the ground truth. With the increase of spraying resolution (decrease of grid size), the spraying rate difference gap between the prescription map and GTM is becoming smaller, from 3.18% at 100×100 grid size to 1.07% at 10×10 grid size. Figure 8 shows the linear relationship ($R^2$=0.949) between herbicide saving rate and grid size based on GTM and Figure 9 shows this linear relationship ($R^2$=0.947) based on the predicted prescription map. The two fitted lines have the same trend, showing the negative relationship between spraying saving rate and spraying resolution (grid size) under the assumption that each nozzle only covers the whole area of the grid and without overlapped spraying between two neighbour nozzles.

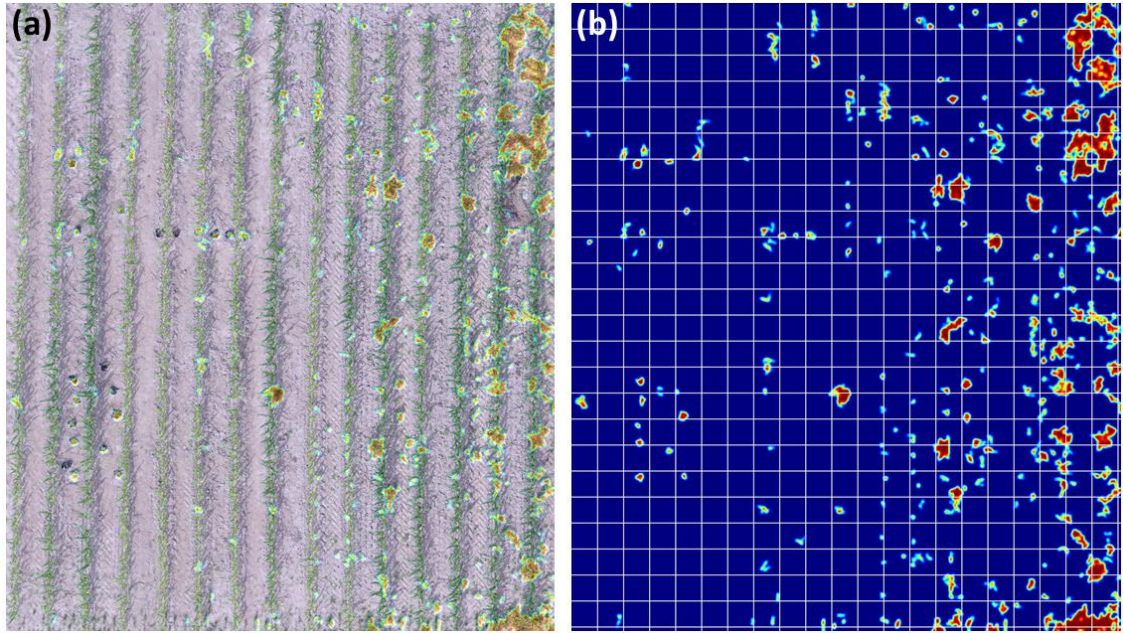

Figure 7. Weed heatmap in the field (a) and spraying prescription map (b) for SSWM (grid size: 100×100 pixels, representing 17.8×17.8 cm$^2$ in the ground).

Table 4. Quantitative results of spot spraying based on two weed maps with three different grid sizes.

| Map* | Free weed grids | Weed grids | Spraying rate | Saving rate |
| --- | --- | --- | --- | --- |
| GTM (100×100) | 254 | 250 | 49.60% | 50.40% |
| Prescription Map (100×100) | 238 | 266 | 52.78% | 47.22% |
| GTM (50×50) | 1466 | 550 | 27.28% | 72.72% |
| Prescription Map (50×50) | 1429 | 587 | 29.12% | 70.88% |
| GTM (10×10) | 46771 | 4080 | 8.02% | 91.98% |
| Prescription Map (10×10) | 46230 | 4621 | 9.09% | 90.91% |

*The grid sizes at 100×100, 50×50, 10×10 represent 17.8×17.8 cm$^2$, 8.9×8.9 cm$^2$ and 1.78×1.78 cm$^2$ ground area, respectively.

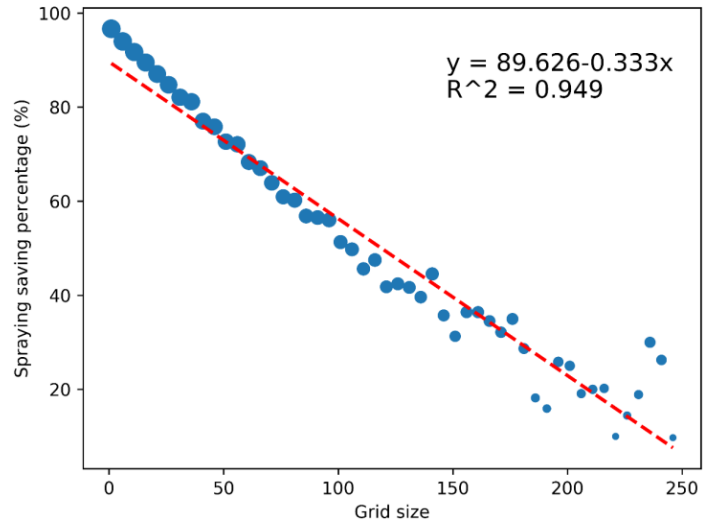

Figure 8. The fitted linear relationship between saved spraying rate and grid size based on the GTM.

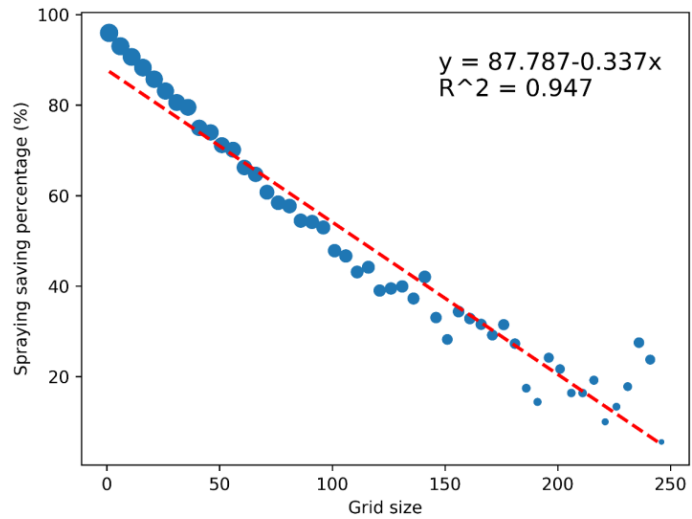

Figure 9. The fitted linear relationship between saved spraying rate and grid size based on the predicted map by the proposed network.

**Feature map visualization**

The developed network trained with field images not only works well in the test field dataset, but also is capable of predicting classes in the remotely sensed UAV images that are never seen by the networks. The outputs of the hidden layers in the network are helpful to understand how a succession of convolutional layers translate their inputs, and to get an intuition about the internal representations for each individual convolutional filter. Figure 10 shows the input image and its feature maps at the first (shallow) and $26^{th}$ (deep) layers. An input image was fed into the proposed network. It can be seen that the outputs in the shallow layers (Conv1) highlight different areas like vegetation and soil background independently within the given image. The content (maize and weed objects) of images can still be distinguished which indicates that the network learns some low-level features such as color, edges, and texture in the shallow layers (512×512×64). There are some feature maps highlighting the maize and weed pixels with the red color. It can be seen that the vegetation part and soil background are displayed with different colors, indicating the shallow layers already extract useful features to segment the vegetation including weeds and maize from the soil background, but not yet to extract enough features to differentiate weed and maize pixels as they show similar color in all feature maps at this layer. In the deeper layer (Conv26), because of a series of convolution, and max pooling operations, the spatial information of feature maps has been eliminated and it is difficult to interpret what features the network learns. The features at this layer (32×32×512) become more abstract. These high-level features are learned but the fine-grained features could be lost because of pooling operations. In order to remediate this issue, we, therefore, concatenated former layers to the deeper layers in the decoder part.

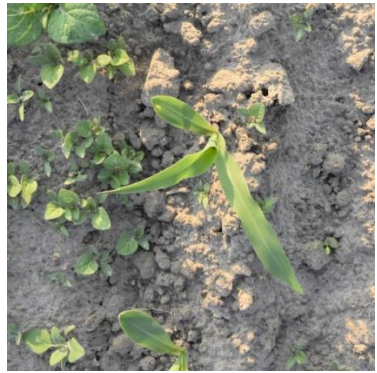

(a)

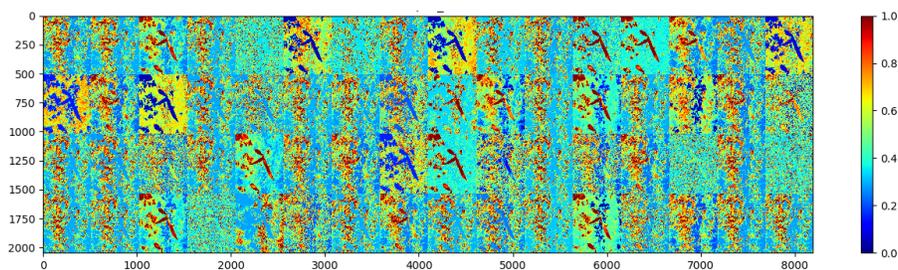

(b)

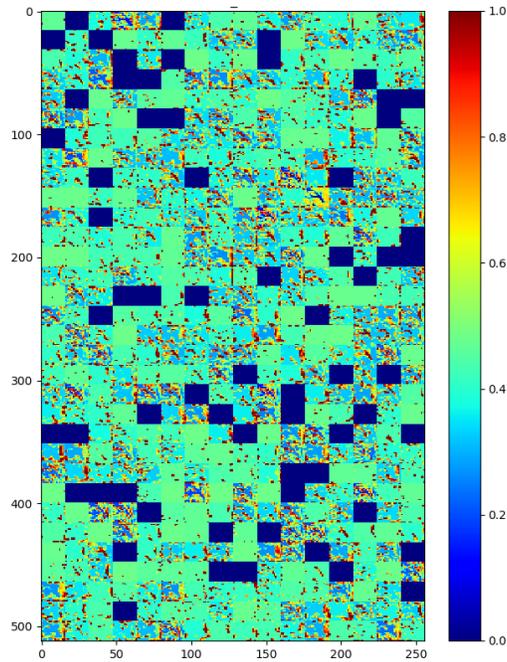

(c)

Figure 10. Input image (a) and its feature maps at the first layer (b) and the twenty sixth layer (c).

## 4. Discussion

The results demonstrate the feasibility of using one single deep neural network trained with field data to produce a drone-based weed map for precision farming. The network achieved satisfactory results both for field data and aerial data. The reasons for the good performance of our model are as follows. First, the use of the max-pooling layer with indices and concatenation in the network largely keeps the spatial information and fine-grained features in the deep layers, which can effectively detect small-sized weeds. Second, the weighted loss function solves the problem of imbalanced samples. This means that the network treats each class equally during the processing of training. The post-processing algorithm (CRFs) can further fine-tune the predicted mask such as noisy small object elimination and object boundary correction. Most importantly, the preprocessing was applied in the field dataset before training to close the gap between the two dataset differences. It is difficult and unfair to directly compare the performances with other studies as different datasets are used. In the previous study, Kim and Park, (2022) developed an MTS-CNN for weed and crop detection and achieved 0.8823 mIOU. This is higher than our proposed network, which achieved mIOU values in our field dataset of 0.767 and aerial dataset of 0.617. The use of MTS-CNN for weed mapping in aerial images has not been explored and validated. The accuracy of the weed map could be improved by adding labelled aerial images with the same semantic objects in the training dataset. Besides, the public open datasets can be used, e.g. in field annotation dataset for sugar beet and weed segmentation. Chebrolu et al., (2017) published field multi-channel image annotation dataset. Sa et al., (2018b) released a large sugar beet and weed aerial annotation dataset. With the popularity of drone-based remote sensing in precision farming, more available open datasets will become available. Thus, transforming knowledge from different datasets is becoming increasingly important to boost the performance of deep neural networks and reduce the labour cost and time of labelling of image data. Typically, crop/weed maps are produced by using part of training samples in the orthomosaic imagery. This procedure inevitably involves manual labelling. Our methodology provides a new direction for weed mapping with UAVs by utilizing other data sources collected with field platforms with the same semantic objects. There are more publicly available field datasets (Chebrolu et al., 2017) for weed and crop segmentation, which could allow transferring the learned features in these datasets to predict and map weeds with UAVs provided the two datasets share the same semantic objects.

The predicted prescription map could be directly integrated with a tractor platform for spot spraying if georeference information is provided. The two fitted lines from GMP (Fig. 8) and the predicted prescription map (Fig. 9) are very close, validating the effectiveness of our network to produce a weed map from UAVs. It is worth noting the failures in the predictions as illustrated in Fig.11, and it is clear that the weed seedlings, especially grass weeds, are difficult objects to segment. Overlapping between objects could be another scenario where the network does not predict well. In aerial imagery, the lower resolution and heavier overlaps could limit the accuracy of weed maps generated from the network. Intuitively, flying at a lower height leading to higher resolutions could help. Other potential methods can be considered by using more advanced imaging sensors such as multispectral or hyperspectral cameras (Gao et al., 2018b), which could utilize spectral features beyond RGB bands. On the other hand, these multispectral data typically have lower resolution (1-2 cm versus millimetre level), which is again making things more difficult to classify. Also, super-resolution imaging, a class of techniques to enhance the resolution of an imaging system, is a way to improve the resolution of aerial images and thus further close the difference gap (Chen et al., 2022).

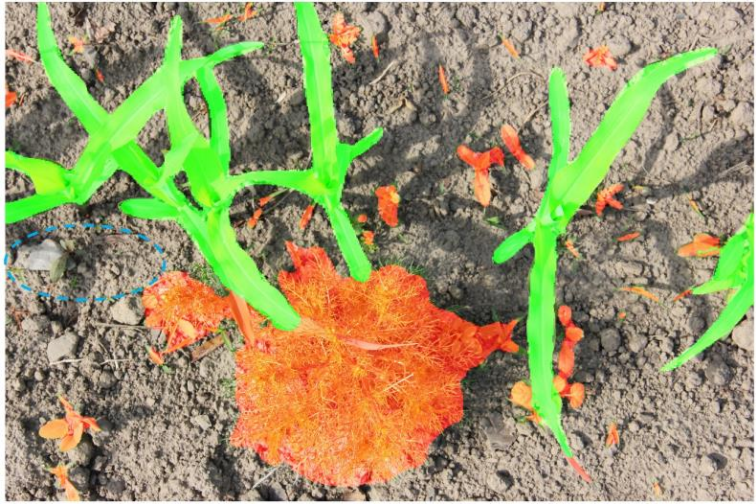

Figure 11. Segmentation result of a field image.

We show the feature learning process of plant objects with feature map visualization. The visualization is helpful to understand what the deep neural network learns. Moreover, it improves network architecture such as model selection and parameter reduction. In Fig.10 (c), we find that some feature maps at deep layers displayed in blue color are not activated and can be regarded as redundant features. This means the network can be further optimized to improve its architecture. The specific relationship between the number of reduction parameters and visualization results is a meaningful work to quantify interpretable parameters of deep neural networks (Toda and Okura, 2019).

## 5. Conclusions

Deep learning has become increasingly significant in semantic segmentation of UAV-based remote sensing imagery and has shown to be powerful to extract multi-level features and find potential patterns from a big image dataset. This paper presents a complete pipeline for semantic weed/crop segmentation using deep convolutional neural networks. The networks were trained only with field images, and then were validated both using field test data and remotely sensed images from a UAV platform. The proposed network outperformed in the field test dataset, shown in Table 2. For the validation of UAV imagery, we first generated an orthomosaic photo from the overlapped aerial images, and then used the network to predict cropped subimages from the orthomosaic photo. Finally, the predictions of subimages were merged to generate crop map. The result in the UAV dataset shows that the proposed network still performs best compared to the other networks.

The use of a weighted loss function is a good way to remediate the issue of imbalanced classes in the dataset. The weights of each class can be quantified with the pixel statistics in the training dataset. Fully connected CRFs were used to fine-tune the predictions by the networks.

We demonstrated that it is feasible to use deep convolutional neural networks trained with field data to predict aerial images based on high-resolution UAV orthomosaic imagery and field images. The crop/weed map finally could be produced through our proposed pipeline, which opens a new path to generate a workflow for site-specific weed management (SSWM) studies. Moreover, the study shows that deep learning has the ability to integrate different source data and can be used for multi-tasks. Our work can benefit some relevant communities such as UAV-based remote sensing, precision crop farming and agricultural field robots.

**Acknowledgement**

This work was supported by the Lincoln Agri-Robotics Grant (E3) from Research England, and the Nordic Council of Ministers, Copenhagen, Denmark (PPP #6P2 and #6P3). We acknowledge the drone pilot from ILVO for data collection.

**References**

Badrinarayanan, V., Kendall, A., Cipolla, R., 2017. SegNet: A Deep Convolutional Encoder-Decoder Architecture for Image Segmentation. IEEE Trans Pattern Anal Mach Intell 39, 2481–2495. https://doi.org/10.1109/TPAMI.2016.2644615

Bawden, O., Kulk, J., Russell, R., McCool, C., English, A., Dayoub, F., Lehnert, C., Perez, T., 2017. Robot for weed species plant-specific management. J Field Robot 34, 1179–1199. https://doi.org/10.1002/rob.21727

Borra-Serrano, I., Peña, J., Torres-Sánchez, J., Mesas-Carrascosa, F., López-Granados, F., 2015. Spatial Quality Evaluation of Resampled Unmanned Aerial Vehicle-Imagery for Weed Mapping. Sensors 15, 19688–19708. https://doi.org/10.3390/s150819688

Burgos-artizzu, X.P., Ribeiro, A., Guijarro, M., Pajares, G., 2011. Real-time image processing for crop / weed discrimination in maize fields. Comput Electron Agric 75, 337–346. https://doi.org/10.1016/j.compag.2010.12.011

Chebrolu, N., Lottes, P., Schaefer, A., Winterhalter, W., Burgard, W., Stachniss, C., 2017. Agricultural robot dataset for plant classification, localization and mapping on sugar beet fields. International Journal of Robotics Research 36, 1045–1052. https://doi.org/10.1177/0278364917720510

Chen, H., He, X., Qing, L., Wu, Y., Ren, C., Sheriff, R.E., Zhu, C., 2022. Real-world single image super-resolution: A brief review. Information Fusion 79, 124–145. https://doi.org/10.1016/j.inffus.2021.09.005

Chen, L.-C., Papandreou, G., Kokkinos, I., Murphy, K., Yuille, A.L., 2018. DeepLab: Semantic Image Segmentation with Deep Convolutional Nets, Atrous Convolution, and Fully Connected CRFs. IEEE Trans Pattern Anal Mach Intell 40, 834–848. https://doi.org/10.1109/TPAMI.2017.2699184

Dian Bah, M., Hafiane, A., Canals, R., 2018. Deep learning with unsupervised data labeling for weed detection in line crops in UAV images. Remote Sens (Basel) 10. https://doi.org/10.3390/rs10111690

Emmi, L., Gonzalez-de-Santos, P., 2017. Mobile robotics in arable lands: Current state and future trends, in: 2017 European Conference on Mobile Robots (ECMR). pp. 1–6. https://doi.org/10.1109/ECMR.2017.8098694

Gao, J., 2020. An exploration of the use of machine learning techniques for site-specific weed management. Ghent University.

Gao, J., French, A.P., Pound, M.P., He, Y., Pridmore, T.P., Pieters, J.G., 2020. Deep convolutional neural networks for image-based Convolvulus sepium detection in sugar beet fields. Plant Methods 16. https://doi.org/10.1186/s13007-020-00570-z

Gao, J., Liao, W., Nuyttens, D., Lootens, P., Vangeyte, J., Pižurica, A., He, Y., Pieters, J.G., 2018a. Fusion of pixel and object-based features for weed mapping using unmanned aerial vehicle imagery. International Journal of Applied Earth Observation and Geoinformation 67, 43–53. https://doi.org/10.1016/j.jag.2017.12.012

Gao, J., Nuyttens, D., Lootens, P., He, Y., Pieters, J.G., 2018b. Recognising weeds in a maize crop using a random forest machine-learning algorithm and near-infrared snapshot mosaic hyperspectral imagery. Biosyst Eng 170, 39–50. https://doi.org/10.1016/j.biosystemseng.2018.03.006

Gao, J., Westergaard, J.C., Sundmark, E.H.R., Bagge, M., Liljeroth, E., Alexandersson, E., 2021. Automatic late blight lesion recognition and severity quantification based on field imagery of diverse potato genotypes by deep learning. Knowl Based Syst 214, 106723. https://doi.org/10.1016/j.knosys.2020.106723

He, K., Zhang, X., Ren, S., Sun, J., 2015. Delving deep into rectifiers: Surpassing human-level performance on imagenet classification, in: Proceedings of the IEEE International Conference on Computer Vision. pp. 1026–1034. https://doi.org/10.1109/ICCV.2015.123

Huang, J., Zhang, X., Xin, Q., Sun, Y., Zhang, P., 2019. Automatic building extraction from high-resolution aerial images and LiDAR data using gated residual refinement network. ISPRS Journal of Photogrammetry and Remote Sensing 151, 91–105. https://doi.org/10.1016/j.isprsjprs.2019.02.019

Kamilaris, A., Prenafeta-Boldú, F.X., 2018. Deep learning in agriculture: A survey. Comput Electron Agric 147, 70–90. https://doi.org/10.1016/j.compag.2018.02.016

Kim, Y.H., Park, K.R., 2022. MTS-CNN: Multi-task semantic segmentation-convolutional neural network for detecting crops and weeds. Comput Electron Agric 199, 107146. https://doi.org/10.1016/J.COMPAG.2022.107146

Kingma, D.P., Ba, J., 2015. Adam: A Method for Stochastic Optimization, in: International Conference for Learning Representations. pp. 1–15. https://doi.org/http://doi.acm.org.ezproxy.lib.ucf.edu/10.1145/1830483.1830503

Kohli, P., Ladický, L., Torr, P.H.S., 2009. Robust higher order potentials for enforcing label consistency. Int J Comput Vis 82, 302–324. https://doi.org/10.1007/s11263-008-0202-0

Krähenbühl, P., Koltun, V., 2011. Efficient Inference in Fully Connected CRFs with Gaussian Edge Potentials, in: Shawe-Taylor, J., Zemel, R.S., Bartlett, P.L., Pereira, F., Weinberger, K.Q. (Eds.), Advances in Neural Information Processing Systems 24. Curran Associates, Inc., pp. 109–117.

LeCun, Y.A., Bengio, Y., Hinton, G.E., 2015. Deep learning. Nature 521, 436–444. https://doi.org/10.1038/nature14539

Long, J., Shelhamer, E., Darrell, T., 2015. Fully Convolutional Networks for Semantic Segmentation. IEEE Trans Pattern Anal Mach Intell 39, 640–651. https://doi.org/10.1109/TPAMI.2016.2572683

Lottes, P., Behley, J., Milioto, A., Stachniss, C., 2018. Fully Convolutional Networks With Sequential Information for Robust Crop and Weed Detection in Precision Farming. IEEE Robot Autom Lett 3, 2870–2877. https://doi.org/10.1109/LRA.2018.2846289

Milioto, A., Lottes, P., Stachniss, C., 2017. REAL-TIME BLOB-WISE SUGAR BEETS VS WEEDS CLASSIFICATION for MONITORING FIELDS USING CONVOLUTIONAL NEURAL NETWORKS, in: ISPRS Annals of the Photogrammetry, Remote Sensing and Spatial Information Sciences. pp. 41–48. https://doi.org/10.5194/isprs-annals-IV-2-W3-41-2017

Noh, H., Hong, S., Han, B., 2015. Learning deconvolution network for semantic segmentation, in: Proceedings of the IEEE International Conference on Computer Vision. pp. 1520–1528. https://doi.org/10.1109/ICCV.2015.178

Oerke, E.C., 2006. Crop losses to pests. J Agric Sci 144, 31–43.

Peerbhay, K., Mutanga, O., Lottering, R., Bangamwabo, V., Ismail, R., 2016. Detecting bugweed (Solanum mauritianum) abundance in plantation forestry using multisource remote sensing. ISPRS Journal of Photogrammetry and Remote Sensing 121, 167–176. https://doi.org/10.1016/j.isprsjprs.2016.09.014

Perez-Ortiz, M., Pena, J.M., Gutierrez, P.A., Torres-Sanchez, J., Hervas-Martinez, C., Lopez-Granados, F., 2016. Selecting patterns and features for between- and within- crop-row weed mapping using UAV-imagery. Expert Syst Appl 47, 85–94. https://doi.org/10.1016/j.eswa.2015.10.043

Sa, I., Chen, Z., Popović, M., Khanna, R., Liebisch, F., Nieto, J., Siegwart, R., 2018a. weedNet: Dense Semantic Weed Classification Using Multispectral Images and MAV for Smart Farming. IEEE Robot Autom Lett 3, 588–595. https://doi.org/10.1109/LRA.2017.2774979

Sa, I., Popović, M., Khanna, R., Chen, Z., Lottes, P., Liebisch, F., Nieto, J., Stachniss, C., Walter, A., Siegwart, R., 2018b. WeedMap: A large-scale semantic weed mapping framework using aerial multispectral imaging and deep neural network for precision farming. Remote Sens (Basel) 10. https://doi.org/10.3390/rs10091423

Shapira, U., Herrmann, I., Karnieli, A., Bonfil, D.J., 2013. Field spectroscopy for weed detection in wheat and chickpea fields. Int J Remote Sens 34, 6094–6108. https://doi.org/10.1080/01431161.2013.793860

Shaw, D.R., 2005. Remote sensing and site-specific weed management. Front Ecol Environ 3, 526–532. https://doi.org/10.1890/1540-9295(2005)003[0526:RSASWM]2.0.CO;2

Shotton, J., Winn, J., Rother, C., Criminisi, A., 2009. TextonBoost for image understanding: Multi-class object recognition and segmentation by jointly modeling texture, layout, and context. Int J Comput Vis 81, 2–23. https://doi.org/10.1007/s11263-007-0109-1

Singh, A.K., Ganapathysubramanian, B., Sarkar, S., Singh, A., 2018. Deep Learning for Plant Stress Phenotyping: Trends and Future Perspectives. Trends Plant Sci. https://doi.org/10.1016/j.tplants.2018.07.004

Thoma, M., 2016. A Survey of Semantic Segmentation. arXiv preprint arXiv:1602.06541.

Toda, Y., Okura, F., 2019. How Convolutional Neural Networks Diagnose Plant Disease. Plant Phenomics 2019, 9237136. https://doi.org/10.1155/2019/9237136

Torres-Sánchez, J., López-Granados, F., Peña, J.M., 2015. An automatic object-based method for optimal thresholding in UAV images: Application for vegetation detection in herbaceous crops. Comput Electron Agric 114, 43–52. https://doi.org/10.1016/j.compag.2015.03.019

van de Vijver, R., Mertens, K., Heungens, K., Somers, B., Nuyttens, D., Borra-Serrano, I., Lootens, P., Roldán-Ruiz, I., Vangeyte, J., Saeys, W., 2020. In-field detection of Alternaria solani in potato crops using hyperspectral imaging. Comput Electron Agric 168, 105106. https://doi.org/10.1016/j.compag.2019.105106

Wang, A., Zhang, W., Wei, X., 2019. A review on weed detection using ground-based machine vision and image processing techniques. Comput Electron Agric 158, 226–240. https://doi.org/10.1016/j.compag.2019.02.005

Wang, N., Zhang, N., Wei, J., Stoll, Q., Peterson, D.E., 2007. A real-time, embedded, weed-detection system for use in wheat fields. Biosyst Eng 98, 276–285. https://doi.org/10.1016/j.biosystemseng.2007.08.007

Zhang, N., Wang, M., Wang, N., 2002. Precision agriculture - A worldwide overview, in: Computers and Electronics in Agriculture. pp. 113–132. https://doi.org/10.1016/S0168-1699(02)00096-0

Zhang, Y., Gao, J., Cen, H., Lu, Y., Yu, X., He, Y., Pieters, J.G., 2019. Automated spectral feature extraction from hyperspectral images to differentiate weedy rice and barnyard grass from a rice crop. Comput Electron Agric 159, 42–49. https://doi.org/10.1016/j.compag.2019.02.018